\documentclass[runningheads]{llncs}
\usepackage[T1]{fontenc}
\usepackage{amsmath}
\usepackage{graphicx}
\usepackage{xurl}
\usepackage[hidelinks]{hyperref}
\usepackage{booktabs}
\begin{document}
\title{KGCQual: An Interpretable Framework for Evaluating the Knowledge Graph Construction Quality from Text}
\titlerunning{KGCQual}
%
 \author{
 Nipun Misra\inst{1} \and Vikranth Udandarao\inst{2}\and
 Aanchal Gupta\inst{2} \and
 Yogender Kumar\inst{2}\and
 Manuj Mukherjee\inst{2}\and
 Raghava Mutharaju\inst{3}
}

 \authorrunning{N. Misra et al.}

\institute{
$^{1}$VIT Vellore, Vellore, India \\
\email{nipun.misra2022@vitstudent.ac.in} \\
$^{2}$IIIT-Delhi, Delhi, India \\
\email{\{aanchal21224, yogender21505, vikranth22570, manuj\}@iiitd.ac.in} \\
$^{3}$Indian Institute of Technology Palakkad, Kerala, India. \\ 
\email{raghava@iitpkd.ac.in}
}


%
\maketitle              
\begin{abstract}
Knowledge Graphs (KGs) are increasingly constructed through automated extraction pipelines; however, such systems often introduce spurious or incomplete triples, which degrade downstream performance. Existing evaluation practices rely heavily on task-specific metrics or small-scale manual verification, offering limited insight into the structural and semantic fidelity of extracted graphs. We propose a novel, interpretable metric for intrinsic KG quality assessment that measures how closely an automatically extracted graph approximates an “ideal” graph capturing the key noun phrases, predicate relations, and basic linguistic phenomena such as negation expressed in the source text. Our framework integrates two complementary components: (1) an entity-level assessment that evaluates completeness, resolution quality, and connectivity, and (2) a relation-level assessment that judges predicate preservation and multiplicity using lexical similarity, dependency-parse alignment, and light-weight negation handling to ensure semantic faithfulness. We evaluate our metric across multiple state-of-the-art triple extraction systems and datasets, including WebNLG, TinyButMighty, and BenchIE, demonstrating that it reliably identifies omissions, redundancy, and structural deviations that existing metrics overlook. Our work offers a scalable, model-agnostic, and interpretable framework for comparing automated KG construction methods and provides a foundation for standardised evaluation. We further validate the metric through an ablation study isolating noun and
verb components, and a downstream evaluation showing that KGCQual scores correlate significantly with link prediction performance ($\rho=-0.900$, $p=0.037$) on the same extracted KGs. The code repository is available at \url{https://github.com/kracr/kg-quality-metric}. 


\end{abstract}

\section{Introduction}

Knowledge Graphs (KGs) have become a foundational abstraction enabling structured representation of entities, relations, and facts in a 
form amenable to reasoning, querying, and integration across heterogeneous data 
sources~\cite{ref_article11,ref_article15}. A growing fraction 
of modern KGs spanning domains such as scientific literature, news, and enterprise documents are constructed through automated Information Extraction (IE) pipelines. These systems transform natural language text into 
subject–predicate–object triples that are later serialized into RDF, linked to 
ontologies, or validated through SHACL constraints. 

However, despite their widespread adoption, the quality of triples produced by 
IE systems remains highly variable. Existing KG quality frameworks 
(e.g.,~\cite{ref_article15,ref_article12}) provide rich dimensions such as accuracy, completeness, and consistency, but \emph{presuppose} that the underlying triples are already correct or human-curated. In contrast, for automatically extracted KGs, the fundamental bottleneck lies one level earlier: the lack of an \emph{intrinsic, sentence-level method} for assessing whether extracted triples faithfully preserve the semantics expressed in the source text. Current evaluations rely 
predominantly on (i) downstream task performance, which is indirect and application-specific, or (ii) manual annotation, which is costly and non-scalable. This creates a persistent gap between Semantic Web quality frameworks (which expect high-integrity RDF graphs) and the practical realities of pipeline-driven KG construction. In particular, no existing metric jointly captures \emph{structural}, \emph{lexical}, and \emph{semantic} fidelity of extracted triples with respect to their textual origin. Yet such intrinsic valuation is critical: errors such as missing entities, merged noun phrases, incorrect relation multiplicity, or undetected negation propagate into downstream KGs and ultimately compromise reasoning, linking, and query results.

To address this gap, we propose \textbf{KGCQual}, a principled and interpretable 
framework for evaluating the quality of automatically constructed triples prior 
to their integration into a KG. Our approach is grounded in the notion of an 
\emph{ideal reference graph} for a sentence - an abstract representation of all 
its salient entities and predicates. While this ideal graph is not observed 
directly, we approximate it by combining part-of-speech cues, dependency 
relations, and lightweight semantic similarity methods. KGCQual evaluates 
triple quality along two complementary dimensions:

\begin{enumerate}
    \item \textbf{Entity Quality (Noun-Level)}: measuring completeness, 
    granularity (resolution), and connectivity of entity references derived 
    from the text. This captures whether the extracted triples maintain the 
    structure expected of RDF nodes corresponding to noun phrases.

    \item \textbf{Relation Quality (Verb-Level)}: assessing predicate accuracy, 
    multiplicity, and semantic similarity, including cases where paraphrasing 
    or morphological variation occurs. This reflects whether the extracted 
    predicates preserve the relational semantics intended in the source 
    sentence.
\end{enumerate}

Together, these form a normalized, interpretable score indicating how well an 
IE tool preserves sentence-level semantics before the triples are serialized 
into RDF or integrated into a broader KG. To illustrate the challenge, consider the following sentence:

\begin{quote}
\textit{``The Urchin Software Configuration Management system is a free and 
open source initiative to provide an enterprise-scale solution to managing the 
configuration of many software applications running on many servers across an 
enterprise.''}
\end{quote}

\begin{figure}[htbp]
    \centering
    \begin{minipage}[t]{0.48\textwidth}
        \centering
        \includegraphics[width=\textwidth]{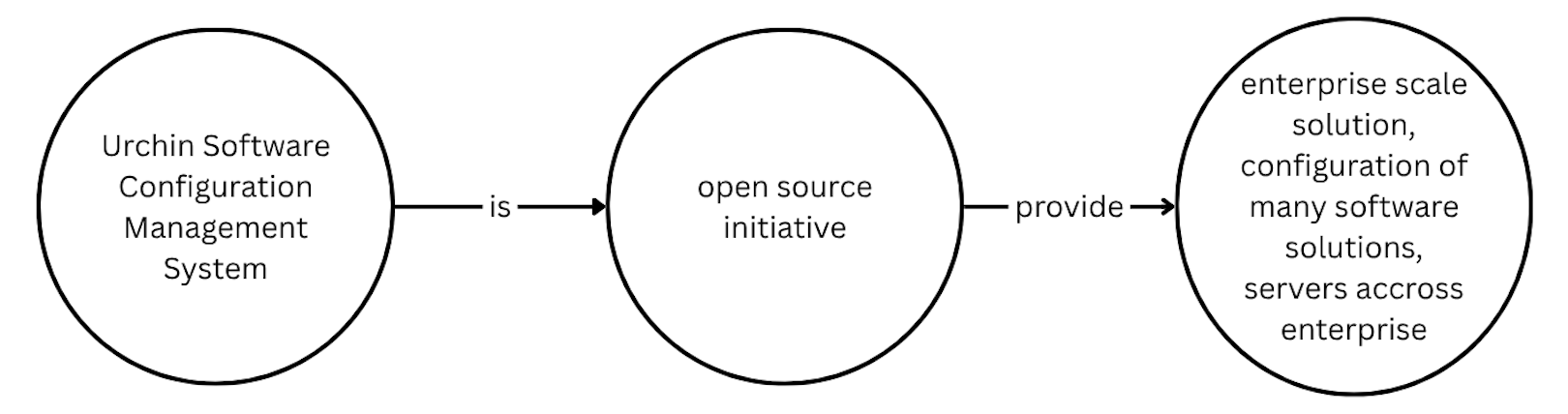}
        \caption{Triple graph extracted by MinIE}
        \label{fig:minie}
    \end{minipage}%
    \hfill
    \begin{minipage}[t]{0.48\textwidth}
        \centering
        \includegraphics[width=\textwidth]{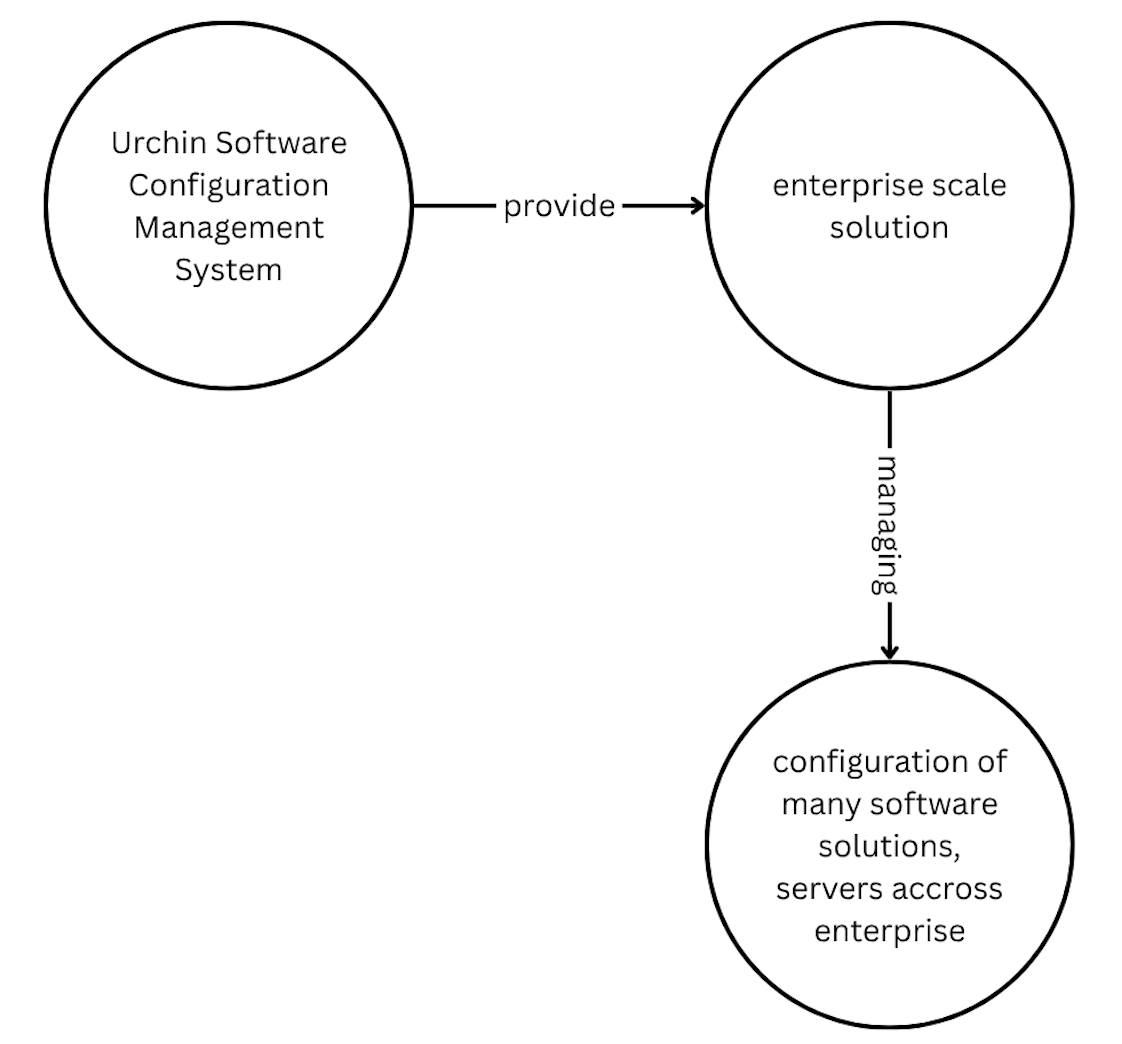}
        \caption{Triple graph extracted by OllIE}
        \label{fig:ollie}
    \end{minipage}
\end{figure}

\begin{figure}[htbp]
    \centering
    \begin{minipage}[t]{0.48\textwidth}
        \centering
        \includegraphics[width=\textwidth]{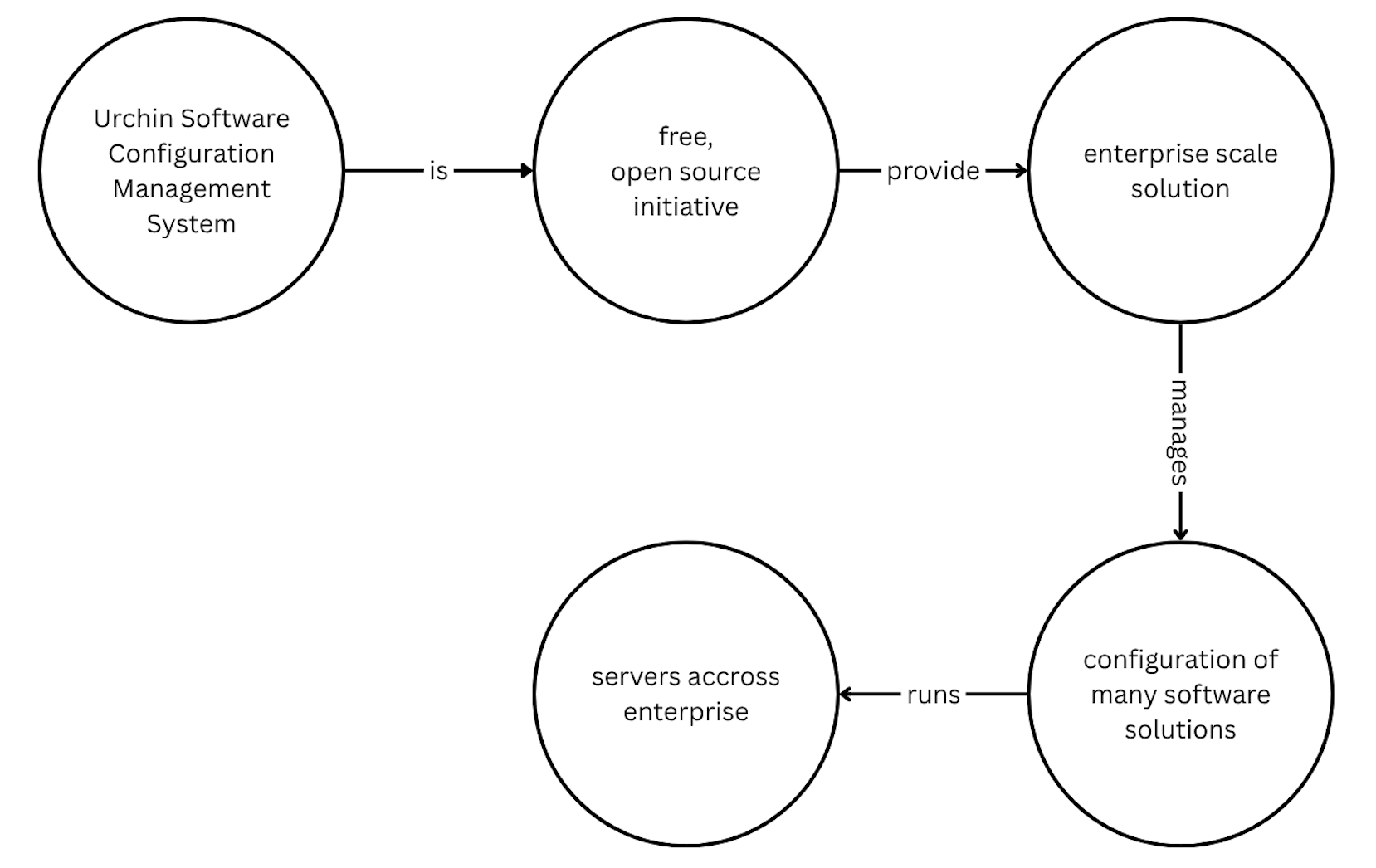}
        \caption{Triple graph extracted by GPT based model}
        \label{fig:gpt}
    \end{minipage}%
    \hfill
    \begin{minipage}[t]{0.48\textwidth}
        \centering
        \includegraphics[width=\textwidth]{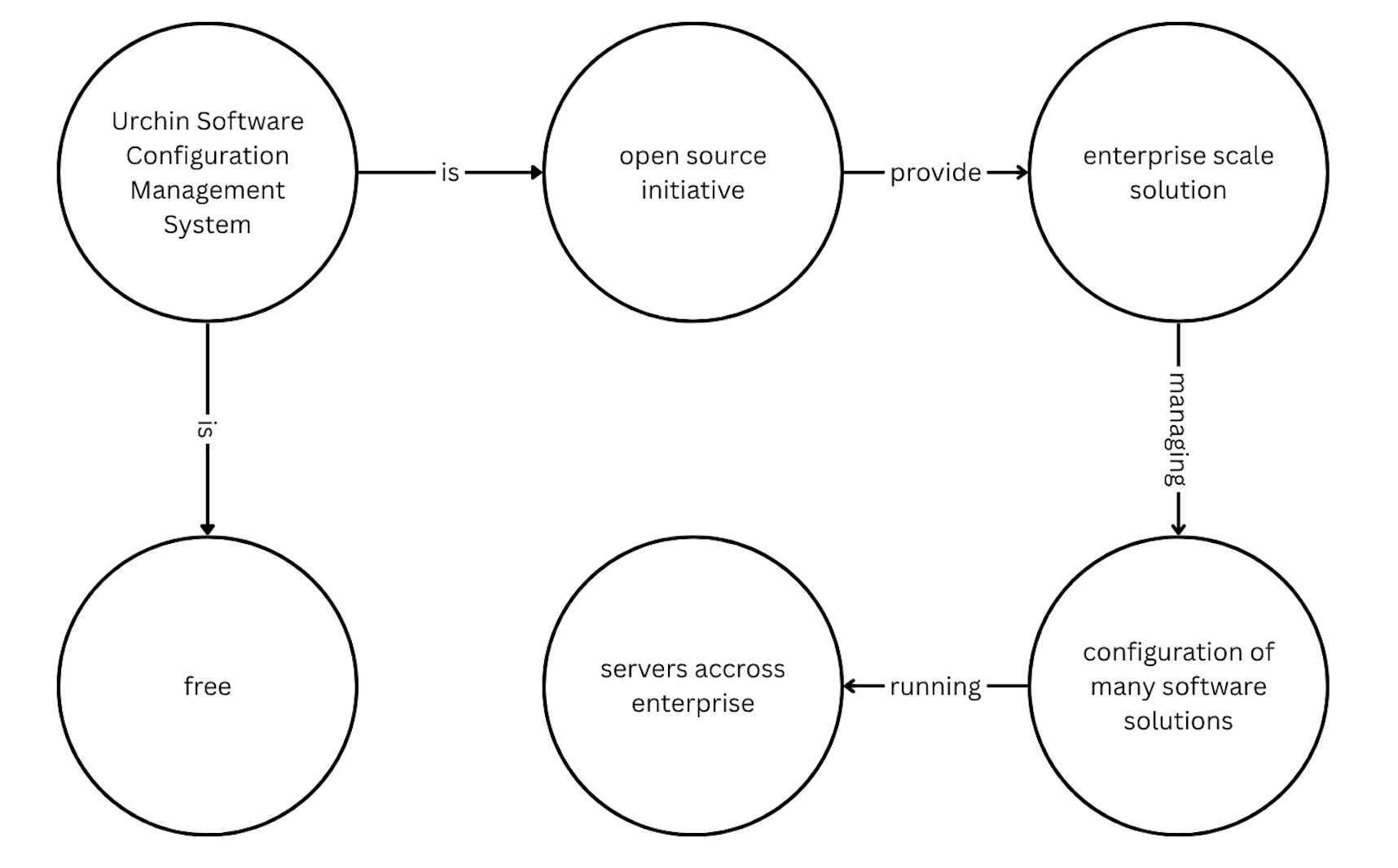}
        \caption{Ideal triple graph derived using POS tagging}
        \label{fig:ideal}
    \end{minipage}
\end{figure}

State-of-the-art IE systems such as MinIE~\cite{ref_article21}, 
OllIE~\cite{ref_article23}, and Large Language Model (LLM)–based extractors 
often omit essential entities (e.g., \textit{``free, open-source initiative''}), 
merge distinct noun phrases (e.g., \textit{``configuration of many software 
applications''}), or reduce verb multiplicity (e.g., collapsing multiple 
\textit{``is''} relations). Figures~\ref{fig:minie}--\ref{fig:ideal} visualize 
these discrepancies. Such errors degrade KG completeness and semantic fidelity 
if not detected early.

KGCQual directly targets these shortcomings by providing a systematic way to 
quantify deviations from an ideal graph, enabling fair comparison of IE tools, 
LLMs, and extraction pipelines. By grounding evaluation in both linguistic 
structures and Semantic Web quality principles, our framework bridges the gap 
between NLP-driven extraction and KG-centric validation.

\section{Related Work}

\subsection{Knowledge Graph Quality Assessment}
Ensuring the quality of Knowledge Graphs (KGs) is very important for downstream applications. Foundational surveys~\cite{ref_article15,ref_article11} outline core dimensions such as accuracy, completeness, consistency, and relevancy, and provide lifecycle-oriented frameworks for evaluating KGs from creation to maintenance. These approaches 
primarily operate \emph{post hoc} on already constructed graphs, identifying conflicting triples, 
schema violations, outdated assertions, or structural anomalies through rule-based checks, 
constraint validation (e.g., SHACL), or statistical consistency measures~\cite{ref_article12}.  
Such methods presuppose that triples ingested into a KG are semantically reliable. However, when 
triples originate from automatic extraction pipelines, the primary source of error lies \emph{prior} 
to graph construction: the extraction stage itself.

\subsection{Triple-Level Intrinsic Evaluation}
Despite the centrality of triples as the atomic units of RDF graphs, intrinsic triple-level 
evaluation remains underexplored. Existing benchmarks such as WebNLG, ReFinED, and BenchIE 
\cite{ref_article11,ref_article12,ref_article15} evaluate IE systems through manually curated gold 
triples or downstream tasks, but they do not offer a \emph{sentence-grounded, interpretable metric} 
that quantifies how well extracted triples preserve the fine-grained semantic structure of the 
source text. Sampling-based accuracy estimation~\cite{ref_article7} helps assess large KGs, but is 
not tailored to decomposing extraction errors in terms of entity completeness, predicate fidelity, 
or structural correspondence to linguistic cues.

In contrast, our work focuses on the earliest stage of KG construction by providing a 
\textit{sentence-level intrinsic metric} that characterizes how faithfully an IE system captures the 
entities and relations expressed in text. This complements existing KG quality frameworks by 
addressing a stage they do not directly evaluate: \emph{the correctness of triples before they 
become part of an RDF graph}.

\subsection{Information Extraction Quality and Semantic Fidelity}
Research in Information Extraction has examined extraction errors such as missing entities, 
incorrect argument boundaries, and predicate omissions. Systems such as MinIE and OllIE 
\cite{ref_article21,ref_article23} reduce syntactic complexity but often at the expense of semantic 
completeness. LLM-based extractors~\cite{ref_article25} improve fluency but may alter predicate forms or merge multi-word noun phrases, reducing the semantic resolution required for high-quality KG.

Several works scrutinize semantic fidelity in greater depth. Negation handling has been shown to 
be a major failure point: recent studies report that state-of-the-art open IE systems incorrectly 
preserve negation in up to 40\% of cases~\cite{ref_article28}, while parse-driven approaches achieve 
improved but still imperfect performance~\cite{ref_article29}. However, these efforts treat negation 
as an isolated linguistic phenomenon rather than integrating it into a broader KG quality model. 
Similarly, semantic similarity measures and paraphrase detection are often used heuristically in 
open IE, yet they are rarely incorporated into principled KG-oriented evaluation metrics.

\subsection{Positioning of Our Approach}
Existing work offers rich frameworks for \emph{graph-level} KG quality and localized techniques for 
\emph{linguistic error detection}, but no unified metric that connects the two. Our contribution 
fills this gap by integrating:

\begin{itemize}
    \item \textbf{Structural fidelity}: completeness, granularity, and connectivity of entity nodes,  
    \item \textbf{Predicate fidelity}: relation multiplicity, lexical/semantic preservation, 
    \item \textbf{Negation fidelity}: correctness of polarity propagation in triples,
\end{itemize}

Unlike prior approaches that evaluate IE systems indirectly through task performance or manually 
annotated datasets, our framework provides a \emph{direct, intrinsic, and interpretable} measure of 
triple correctness with respect to the source sentence. This positions our work as a bridge between 
linguistically grounded IE evaluation and Semantic Web quality assurance, addressing a critical but 
largely unexamined stage of the KG lifecycle.

\section{Proposed Methodology}

\subsection{Motivation}
Information Extraction (IE) systems—rule-based, statistical, or LLM-driven—serve as the first step
in automatically constructing Knowledge Graphs (KGs) from natural language text. However, different
extractors often produce triples of varying granularity, completeness, and relational accuracy,
leading to inconsistencies in the resulting KG. Current KG quality frameworks assume that triples
are already correct when ingested; yet in practice, a substantial proportion of KG errors originate
\emph{prior} to integration, during extraction.

We therefore introduce a principled metric for \emph{intrinsic} evaluation of triple extraction
quality. The metric compares an extractor’s output against an \emph{ideal graph} derived directly
from linguistic structure. It decomposes triple quality into two fundamental dimensions:
(1) \textbf{entity fidelity} (noun phrases), and (2) \textbf{relation fidelity} (verbs, predicate
multiplicity). This reflects how well the extracted KG preserves the sentence’s informational
structure before RDF serialization or ontology grounding.

\subsection{Ideal Graph Construction}
For a given sentence, we construct an \emph{ideal reference graph} $G_i$ whose nodes correspond to distinct noun phrases and whose edges correspond to verb phrases derived from
dependency parsing. The ideal graph encodes:
\begin{itemize}
    \item All noun phrases as distinct nodes,
    \item All verb phrases (including multiplicities),
    \item Polarity (negation) when expressed in the sentence,
    \item A single connected component for each sentence.
\end{itemize}

This graph serves as the target structure for quality comparison. The IE system’s output induces a
graph $G_1$, which we evaluate against~$G_i$ using the metrics defined below.

\subsection{Noun Metric}

The noun metric $M_N$ assesses whether the extracted KG preserves the expected set of entities with
respect to:
\begin{enumerate}
    \item \textbf{Completeness} -- all noun phrases appear in $G_1$,
    \item \textbf{Resolution} -- each noun phrase is represented as a distinct node,
    \item \textbf{Connectivity} -- the extracted graph forms a single connected component.
\end{enumerate}

Let $N(G_x)$ be the number of noun phrases, 
$V(G_x)$ be the number of nodes, and $C(G_x)$ be the number of connected components.

\subsubsection{Resolution}
The ideal graph satisfies $V(G_i) = N(G_i)$. Deviation from this equality reflects the merging of
distinct noun phrases:
\[
1 - \frac{V(G_1)}{N(G_1)}.
\]


\subsubsection{Negation-Aware Matching}
To handle polarity-sensitive nouns, we incorporate negation markers into the
string matching process. We detect negation using the \textit{Universal Dependencies} (UD)
syntactic framework, which annotates grammatical relations such as the
\texttt{neg} edge linking a negation word (e.g., ``not'', ``no'') to its head noun or verb.
This allows us to determine whether an extracted noun is negated or not.

Let $\lambda = 0.3$ be a penalty constant and $\text{JW}$ denote
the Jaro--Winkler similarity. For each ideal noun $n_j$:
\[
\tau_{n_j}^{\neg} =
\begin{cases}
\text{JW}(n_j^{\text{ext}}, n_j^{\text{ideal}}), & \text{if their UD-based negation status matches}, \\
\lambda\cdot\text{JW}(n_j^{\text{ext}}, n_j^{\text{ideal}}), & \text{otherwise}.
\end{cases}
\]
$\lambda$ is the weight of the penalty for negations, determining how much emphasis is placed on polarity mismatch penalties. For this study, we fix $\lambda = 0.3$ to assign a reasonable penalty for negation errors without allowing them to overshadow the overall score.
This replaces strict matching with a graded penalty, ensuring that polarity
violations (e.g., \emph{``cat''} vs.\ \emph{``not cat''}) are treated as entity-level errors.


\subsubsection{Completeness}
Missing noun phrases incur additive penalties: \[N(G_i) - N(G_1)\]

\subsubsection{Connectivity}
The ideal graph has a single component, so:
\[
C(G_1) - 1
\]

\subsubsection{Combined Noun Score}
Aggregating all deviations yields:
\[
d_N(G_1,G_i)
=
\left(1 - \frac{V(G_1)}{N(G_1)}\right)
+ \sum_{j=1}^{N(G_i)} (1 - \tau_{n_j})
+ (C(G_1) - 1)
\]

The range of this quantity is $[0, 2N(G_i)]$; therefore, we normalize:
\[
M_N = \frac{d_N(G_1,G_i)}{2 N(G_i)}
\]
A value of $0$ indicates perfect noun-level fidelity to the ideal graph.

\subsection{Verb Metric}

Relations (verbs) are central to KG semantics, encoding the predicate structure linking entities.
Errors typically arise in two forms: \emph{loss of multiplicity} and \emph{lexical drift}. The verb
metric captures both.

Let: \[Vb(G_x) = \text{number of predicate instances (including multiplicities)}\]

\subsubsection{Predicate Multiplicity}
Dependency parsing provides reliable subject–verb–object structures and reveals when a verb governs multiple conjunctive objects (e.g., likes cricket, football, and basketball). Missing predicate instances are penalized as:
\[
Vb(G_i) - Vb(G_1)
\]

\subsubsection{Lexical and Semantic Similarity}
We compute a similarity score $\tau_{v_j}$ for each extracted predicate using:
\[
\tau_{v_j} = 
\max\Big(
\text{JW}(v^{\text{ext}}, v^{\text{ideal}}),
\cos(\mathbf{e}(v^{\text{ext}}), \mathbf{e}(v^{\text{ideal}}))
\Big),
\]
where $\mathbf{e}(\cdot)$ denotes contextual or static word embeddings. Lemmatization is applied
prior to similarity computation.

\subsubsection{Combined Verb Score}
The unnormalized verb deviation is:
\[
d_V(G_1,G_i)
=
\left(Vb(G_i) - Vb(G_1)\right)
+ \sum_{j=1}^{Vb(G_i)} (1 - \tau_{v_j})
\]

The normalized verb metric is:
\[
M_V = \frac{d_V(G_1,G_i)}{2Vb(G_i)}
\]

\subsection{Final Metric}
The complete triple-quality metric, KGCQual, combines entity and relation fidelity:
\[
M = \alpha M_N + (1 - \alpha) M_V
\]
where $\alpha \in [0,1]$ controls the relative weight of entity vs. relation quality. We set $\alpha$ to 0.5 to indicate that both the components are equally important. This produces a score in $[0,1]$, where $0$ represents perfect fidelity to the ideal graph. The
metric is extractor-agnostic and can be applied to any pipeline producing subject–predicate–object
triples, enabling principled comparison of heterogeneous IE systems prior to KG integration.

\section{Evaluation}
This section evaluates the proposed metric across eight extraction systems -- five classical OpenIE tools (MinIE~\cite{ref_article21}, OllIE~\cite{ref_article23}, Stanford OpenIE 4.5.3/4.5.6~\cite{ref_article24}, ClausIE~\cite{ref_article22}) and three LLM-based extractors (Claude 3.7~\cite{ref_article26}, Gemini 2.5 Pro~\cite{ref_article27}, GPT-4.0-mini~\cite{ref_article25}). For all experiments, noun phrases are identified using the NLTK POS tagger~\cite{ref_article18}. The evaluation assesses each system’s ability to preserve (i) noun-phrase granularity and coverage, and (ii) predicate multiplicity and semantic accuracy, as captured by our noun and verb metrics.

\subsection{Datasets}
We evaluate on three widely used benchmarks for triple extraction, chosen for their linguistic diversity and annotation quality:

\begin{itemize}
    \item \textbf{WebNLG}~\cite{ref_article9}: 643 English sentences paired with 5408 gold triples (avg.\ 8.41 triples/sentence), covering diverse DBpedia domains.
    \item \textbf{TinyButMighty}~\cite{ref_article10}: 720 complex and compound sentences yielding 3764 annotated triples (avg.\ 5.28 triples/sentence); designed to stress-test OpenIE systems.
    \item \textbf{BenchIE}~\cite{ref_article17}: 300 naturally occurring sentences with 3307 gold triples (avg.\ 11.02 triples/sentence); a challenging testbed for fine-grained relational extraction.
\end{itemize}

These datasets jointly capture a broad range of syntactic constructions, entity configurations, and predicate forms—conditions necessary to evaluate the behaviour of extraction systems under realistic linguistic variability.

\subsection{Experimental Protocol}
For each sentence, we (i) extract triples using each system, (ii) construct the corresponding KG, (iii) compute our noun and verb metrics against the ideal graph obtained via POS tagging and dependency parsing, and (iv) average scores over all sentences of each dataset. Lower scores indicate higher extraction fidelity.

To illustrate metric behaviour, we present two representative sentences, showing how the metric penalises missing noun-phrase resolution, incomplete predicate multiplicity, and semantic deviation.

\subsubsection{Example 1}
\textbf{Input:} \\
\emph{``The asteroid 1147 Stavropolis has an apoapsis of 418476000000.0, a rotation period of 20378.5, a periapsis of 260855000000.0, and an orbital period of 1249.6 days.''}

ClausIE produces a single aggregated triple, collapsing four verb relations. Since the POS tagger detects four predicate instances, the metric heavily penalises loss of multiplicity while rewarding perfect noun coverage. Using the provided counts:
\[
M_N = 0.06, \qquad M_V = 0.375, \qquad M = 0.2175.
\]
This illustrates the intended behaviour: noun extraction is faithful, but predicate structure is under-specified.

\subsubsection{Example 2}
\textbf{Input:} \\
\emph{``The Mason School of Business are the current tenants of Alan B Miller Hall, located at 101 Ukrop Way is in the United States and owned by the College of William and Mary.''}

Stanford OpenIE (v4.5.3) misses one noun phrase and splits the graph into two components. Verb extraction is partially correct but shows lexical mismatch. Computing the metric gives:
\[
M_N = 0.0914, \qquad M_V = 0.1744, \qquad M = 0.1329.
\]
The example demonstrates how the metric simultaneously captures noun loss, graph disconnectivity, and predicate variation.

\subsection{Quantitative Results}

Tables~\ref{tab:metric_results}–\ref{tab:benchie_results} report the average noun metric, verb metric, and final combined score for all systems across all datasets. A lower value denotes higher triple-extraction fidelity.

\begin{table}[htbp]
\caption{Metric values for IE Tools and LLMs on WebNLG dataset}
\centering
\resizebox{\columnwidth}{!}{%
\begin{tabular}{|l|c|c|c|}
\hline
\textbf{IE Tool} & \textbf{Avg Noun Metric} & \textbf{Avg Verb Metric} & \textbf{Final Metric} \\
\hline
MinIE & 0.5632 & 0.5847 & 0.5740 \\
OllIE & 0.3682 & 0.5019 & 0.4351 \\
Stanford 4.5.3 OpenIE & 0.3971 & 0.4143 & 0.4057 \\
Stanford 4.5.6 OpenIE & 0.3795 & 0.4998 & 0.4397 \\
ClausIE & 0.2037 & 0.2415 & 0.2226 \\
Claude 3.7 & 0.3146 & 0.3429 & 0.3287 \\
Gemini 2.5 Pro & 0.2819 & 0.3117 & 0.2968 \\
GPT-4.0-mini & 0.2694 & 0.2841 & 0.2768 \\
\hline
\end{tabular}%
}
\label{tab:metric_results}
\end{table}

\begin{table}[htbp]
\caption{Metric values for IE Tools and LLMs on TinyButMighty dataset}
\centering
\resizebox{\columnwidth}{!}{%
\begin{tabular}{|l|c|c|c|}
\hline
\textbf{IE Tool} & \textbf{Avg Noun Metric} & \textbf{Avg Verb Metric} & \textbf{Final Metric} \\
\hline
MinIE & 0.4558 & 0.5346 & 0.4952 \\
OllIE & 0.3127 & 0.4595 & 0.3861 \\
Stanford 4.5.3 OpenIE & 0.3614 & 0.3863 & 0.3739 \\
Stanford 4.5.6 OpenIE & 0.3115 & 0.4489 & 0.3802 \\
ClausIE & 0.2804 & 0.2281 & 0.2543 \\
Claude 3.7 & 0.2631 & 0.2915 & 0.2773 \\
Gemini 2.5 Pro & 0.2514 & 0.2786 & 0.2650 \\
GPT-4.0-mini & 0.2298 & 0.2432 & 0.2365 \\
\hline
\end{tabular}%
}
\label{tab:tbm_results}
\end{table}

\begin{table}[htbp]
\caption{Metric values for IE Tools and LLMs on BenchIE dataset}
\centering
\resizebox{\columnwidth}{!}{%
\begin{tabular}{|l|c|c|c|}
\hline
\textbf{IE Tool} & \textbf{Avg Noun Metric} & \textbf{Avg Verb Metric} & \textbf{Final Metric} \\
\hline
MinIE & 0.6347 & 0.6784 & 0.6566 \\
OllIE & 0.4536 & 0.6125 & 0.5330 \\
Stanford 4.5.3 OpenIE & 0.3911 & 0.4530 & 0.4221 \\
Stanford 4.5.6 OpenIE & 0.4367 & 0.5872 & 0.5119 \\
ClausIE & 0.2173 & 0.2627 & 0.2400 \\
Claude 3.7 & 0.3147 & 0.3364 & 0.3256 \\
Gemini 2.5 Pro & 0.2893 & 0.3106 & 0.3000 \\
GPT-4.0-mini & 0.2682 & 0.2829 & 0.2756 \\
\hline
\end{tabular}%
}
\label{tab:benchie_results}
\end{table}

\subsection{Gold-Standard Triple Evaluation}
To contextualize system performance, we run our metric on the gold-standard triples in each dataset. As shown in Table~\ref{tab:gold_metrics}, all gold sets achieve substantially lower scores than automated systems:
\[
\text{WebNLG: }0.1585,\quad
\text{TinyButMighty: }0.2486,\quad
\text{BenchIE: }0.1424.
\]
These values establish empirical lower bounds for our metric and confirm alignment with human-annotated extractions.

\begin{table}[htbp]
\caption{Metric values for handwritten triples on the datasets}
\centering
\resizebox{\columnwidth}{!}{%
\begin{tabular}{|l|c|c|c|}
\hline
\textbf{Dataset} & \textbf{Avg Noun Metric} & \textbf{Avg Verb Metric} & \textbf{Final Metric} \\
\hline
WebNLG & 0.1734 & 0.1435 & 0.1585 \\
TinyButMighty & 0.2212 & 0.2760 & 0.2486 \\
BenchIE & 0.1320 & 0.1528 & 0.1424 \\
\hline
\end{tabular}%
}
\label{tab:gold_metrics}
\end{table}

\subsection{Interpretation via Likert-Scale Categorization}
For qualitative interpretability, we map final metric scores to a 5-point Likert scale, anchored by the empirical thresholds observed in gold-standard triples (\(\le 0.25\)). This mapping provides a practical diagnostic for extraction quality.

\begin{itemize}
    \item \textbf{Grade 1 (Excellent):} $M \le 0.25$
    \item \textbf{Grade 2–4 (Intermediate):} $0.25 < M < 0.45$
    \item \textbf{Grade 5 (Poor):} $M \ge 0.45$
\end{itemize}

Across all datasets, ClausIE consistently falls into Grade~1, while MinIE occupies Grade~5. LLM-based extractors outperform traditional OpenIE systems on average but do not match gold-standard quality.

This analysis highlights that the metric discriminates reliably across systems with varying extraction behaviour and correlates well with qualitative judgments.

\section{Discussion}

This section examines the diagnostic and discriminative capabilities of the proposed metric by analyzing error patterns exhibited by two representative extraction systems: ClausIE and Stanford CoreNLP 4.5.6. We select sentences yielding extremely low and high metric scores and conduct detailed case studies. The goal is to demonstrate that the metric not only quantifies extraction quality but also functions as an interpretable tool for identifying systematic weaknesses in IE pipelines.

\subsection{Tool-Level Behaviour}

Across all datasets, ClausIE exhibits substantial variance in extraction quality. For instance, the following sentence receives a near-optimal score ($\approx 0.006$):

\begin{quote}
\emph{Cornell University is in Ithaca, New York and their nickname is Cornell Big Red. They are the publisher of Administrative Science Quarterly and are affiliated with the Association of American Universities.}
\end{quote}

Here ClausIE successfully preserves both noun-phrase structure and predicate multiplicity. In contrast, it assigns a much higher score ($0.392$) to:

\begin{quote}
\emph{Angola International Airport is located in Ícolo e Bengo, Luanda Province, Angola. The runway is 4000ft long and is 159m a.s.l.}
\end{quote}

reflecting noun-phrase fusion, predicate under-specification, and incomplete coverage of relational structure. Stanford CoreNLP 4.5.6 shows a similar pattern. A low score ($\approx 0.037$) is obtained for:

\begin{quote}
\emph{Alan Shepard has died in California. He was born in New Hampshire and graduated from NWC MA in 1957. He served as a test pilot.}
\end{quote}

while a more linguistically complex example:

\begin{quote}
\emph{Above the Veil is an Australian novel and the sequel to Aenir and Castle. It was followed by Into Battle and The Violet Keystone.}
\end{quote}

yields a substantially higher score ($0.446$). This variation demonstrates that the metric is sensitive to structural, lexical, and contextual extraction failures.

\subsection{Case Study 1: ClausIE}

Consider:

\begin{quote}
\emph{The inaugural Principal was Mr Reidal who managed the school from 1953 to 1956, the next principal was Bill Walker who served from 1957 to 1959.}
\end{quote}

ClausIE outputs:
\begin{itemize}
    \item \texttt{[the next principal, was, Bill]}
    \item \texttt{[The inaugural Principal, was, Mr Reidal who managed school from 1953]}
\end{itemize}

The score ($> 0.3$) reflects several extraction defects:

\begin{itemize}
    \item \textbf{Noun incompleteness}: spans such as \texttt{1957 to 1959} and the full temporal phrase \texttt{school from 1953 to 1956} are truncated or omitted.
    \item \textbf{Low resolution}: compound noun phrases are merged, preventing correct KG structure from emerging.
    \item \textbf{Predicate omission}: \texttt{managed} and \texttt{served} are missing, underspecifying the event structure.
\end{itemize}

The ideal noun and verb sets, derived via POS tagging and dependency parsing, are:

\begin{itemize}
    \item \textbf{Nouns:} \texttt{inaugural Principal, Mr Reidal, school from 1953 to 1956, next principal, Bill Walker, 1957 to 1959}
    \item \textbf{Verbs:} \texttt{was, managed, was, served}
\end{itemize}

Corrected triples:

\begin{itemize}
    \item \texttt{[The inaugural Principal, was, Mr Reidal]}
    \item \texttt{[Mr Reidal, managed, the school from 1953 to 1956]}
    \item \texttt{[The next principal, was, Bill Walker]}
    \item \texttt{[Bill Walker, served, from 1957 to 1959]}
\end{itemize}

Evaluated with our metric, these revised triples yield a reduced score ($0.045$), demonstrating that the metric rewards improvements aligned with KG quality principles such as granularity, coverage, and relation fidelity.

\subsection{Case Study 2: Negation Handling}

Negation represents an important form of semantic polarity that is often mishandled by extraction tools. For example:

\begin{quote}
\emph{``The pharmaceutical company did not receive FDA approval, nor did it secure investor funding.''}
\end{quote}

ClausIE extracts:
\begin{itemize}
    \item \texttt{(pharmaceutical company, received, FDA approval)}
    \item \texttt{(pharmaceutical company, secured, investor funding)}
\end{itemize}

The baseline metric (without negation) assigns a low score ($0.19$), despite both the triples reversing the meaning of the sentence. Using negation-aware matching, the score becomes $0.58$, a $205\%$ increase. This confirms that the negation extension substantially improves semantic robustness by penalizing polarity-inverting extractions.




\subsection{Insights and Implications}

Across both the case studies, the metric demonstrates three important properties:

\begin{enumerate}
    \item \textbf{Interpretability:} Each metric component maps directly to identifiable extraction behaviours (e.g., noun fusion, predicate loss, disconnectivity, negation omission), offering actionable diagnostics.
    \item \textbf{Sensitivity:} Small structural or lexical refinements in triples consistently produce measurable metric improvements, supporting its use in iterative model development.
    \item \textbf{Semantic grounding:} The negation-aware variant captures distinctions missed by surface-level comparison metrics, improving alignment with KG correctness criteria.
\end{enumerate}

These results show that the metric is not merely a scoring mechanism but a principled evaluative tool that exposes fine-grained extraction failures and guides targeted system improvements.

\subsection{Ablation Study: Noun vs.\ Verb Metric Contribution}
\label{sec:ablation}

To understand the relative contribution of each metric component, we conduct an
ablation study on the TinyButMighty dataset, evaluating three weighting scenarios:
noun-only ($w_N=1.0,\,w_V=0.0$), verb-only ($w_N=0.0,\,w_V=1.0$), and equal weight for nouns and verbs. Table~\ref{tab:ablation} reports the average scores per tool under each scenario. Lower scores indicate higher fidelity.

\begin{table}[htbp]
\caption{Ablation study: average scores under different noun/verb weightings
         on TinyButMighty (lower is better)}
\centering
\resizebox{\columnwidth}{!}{%
\begin{tabular}{|l|c|c|c|}
\hline
\textbf{Tool} & \textbf{Noun Only (1.0, 0.0)} & \textbf{Verb Only (0.0, 1.0)} & \textbf{Equal (0.5, 0.5)} \\
\hline
ClausIE                  & 0.1538 & 0.2301 & 0.1919 \\
MiniE                    & 0.3078 & 0.5712 & 0.4395 \\
Ollie                    & 0.2208 & 0.4657 & 0.3432 \\
Stanford 4.5.3           & 0.2431 & 0.4740 & 0.3586 \\
Stanford 4.5.6           & 0.2475 & 0.6441 & 0.4458 \\
Claude 3.7               & 0.3546 & 0.5443 & 0.4495 \\
Gemini 2.5 Pro           & 0.3674 & 0.5495 & 0.4585 \\
GPT-4o mini              & 0.4043 & 0.6891 & 0.5467 \\
Ideal (TinyButMighty)    & 0.3483 & 0.2760 & 0.3122 \\
\hline
\end{tabular}%
}
\label{tab:ablation}
\end{table}

Three findings emerge. First, \textbf{predicate extraction is the primary
weakness} across all automated tools: every system scores worse on the
verb-only scenario than on the noun-only scenario, confirming that relation
extraction is harder than entity extraction. Second, the \textbf{gold-standard
ideal is the sole exception}—its verb score (0.276) is lower than its noun
score (0.348), indicating that human-annotated triples capture predicates more
faithfully than entities. Third, \textbf{ClausIE achieves the best overall
score} (0.1919 under equal weights), while GPT-4o mini performs the worst (0.5467),
suggesting that LLM-based extractors over-generate or paraphrase predicates in
ways that hurt verb fidelity despite their fluent outputs.

\subsection{Downstream Validation: Correlation with Link Prediction}
\label{sec:downstream}

To validate that KGCQual scores reflect \emph{downstream utility}, we test
whether systems with lower (better) scores also produce KGs that support better
link prediction. For each IE system, we use its extracted triples,
train a Tucker~\cite {balazevic2019tucker} embedding model on it, and measure the 
Mean Reciprocal Rank (MRR) on a test split. Across all the three datasets, Table~\ref{tab:downstream}
shows a near-monotonic correspondence: as KGCQual worsens, MRR falls
consistently. The Spearman rank correlation is $\rho = -0.900$ ($p = 0.037$),
confirming statistical significance.

\begin{table}[t]
\caption{KGCQual scores vs.\ TuckER link-prediction MRR
(aggregated across WebNLG, TinyButMighty, BenchIE;
lower KGCQual = better; higher MRR = better).}
\label{tab:downstream}
\centering
\small
\begin{tabular}{lccc}
\toprule
\textbf{IE System} & \textbf{KGCQual $\downarrow$} & \textbf{MRR $\uparrow$} & \textbf{Hits@10 $\uparrow$} \\
\midrule
ClausIE        & 0.130 & 0.141 & 0.200 \\
Ollie          & 0.207 & 0.107 & 0.157 \\
Stanford 4.5.3 & 0.226 & 0.085 & 0.141 \\
MiniE          & 0.278 & 0.059 & 0.113 \\
Stanford 4.5.6 & 0.360 & 0.073 & 0.144 \\
\midrule
\multicolumn{4}{l}{\textit{Spearman $\rho = -0.900$, $p = 0.037$}} \\
\bottomrule
\end{tabular}
\end{table}

This result provides direct evidence that KGCQual is not merely a structural
proxy but correlates with real downstream task performance. IE systems that
better preserve entity and predicate structure produce KGs on which embedding
models learn more effectively.

\section{Limitations}

Although the proposed noun–verb metric provides a principled, structurally grounded measure of extraction quality, it exhibits important limitations with respect to deeper semantic validation. The metric is intentionally aligned with surface-form properties, such as completeness, resolution, lexical correspondence, and predicate multiplicity, but does not yet assess whether the extracted triples preserve the \emph{intended meaning} or \emph{global semantics} of the sentence. A key failure case emerges when triples are structurally plausible but semantically incoherent. The metric is sensitive to \emph{structural fidelity} but not to \emph{semantic coherence}.

Future work will focus on augmenting the metric with semantic components that go beyond lexical alignment. Planned extensions include:
\begin{itemize}
    \item \textbf{Semantic plausibility checks}: assessing whether a triple aligns with the sentence's propositional content rather than simply containing the correct lexical elements.
    \item \textbf{Contextual semantic validation}: exploring embeddings, entailment models, or sentence–triple alignment methods to evaluate whether extracted relations are semantically faithful to the original text.
\end{itemize}

These additions aim to close the gap between structural correctness and semantic correctness, enabling more comprehensive evaluation of IE systems within Knowledge Graph construction pipelines.


\section{Conclusion}

We introduce KGCQual, a sentence-level intrinsic metric for evaluating the quality of triples extracted by Information Extraction (IE) systems, including LLMs. Our formulation captures two fundamental dimensions of extraction quality relevant to Knowledge Graph (KG) construction: (i) noun-phrase fidelity, reflecting completeness, granularity, and graph connectivity, and (ii) predicate fidelity, capturing verb multiplicity and lexical alignment. Grounded in POS tagging and dependency parsing, the metric provides a transparent, interpretable measure of how closely a system’s extracted triples approximate an ideal sentence-level graph. Our evaluation results show that systems merging noun phrases, omitting key entities, or under-generating predicates score higher (worse), while systems aligning with human-annotated gold triples score lower (better). This confirms the metric behaves as intended and signals extraction fidelity.

An ablation study confirms that predicate extraction is the primary weakness across all systems, with verb-metric scores consistently higher than noun-metric scores. A downstream validation further demonstrates that KGCQual scores correlate significantly with link prediction MRR, providing direct evidence that the metric reflects downstream KG quality and not merely surface-level structural alignment.



By offering a principled, interpretable, and computationally efficient evaluation method, this work provides a practical tool for enhancing the reliability of triple extraction pipelines in Knowledge Graph applications.

\section*{Declaration of use of Generative AI}
    
We used OpenAI's ChatGPT to help polish some sentences for clarity and to assist in formatting parts of this paper in LaTeX. All technical content, implementation, validation, and analysis were developed and verified by the authors, who take full responsibility for the work.

%
%

\bibliographystyle{splncs04}
\bibliography{mybibliography}


%

\appendix

\section{Ablation Study: Full Results}
\label{app:ablation}

Table~\ref{tab:ablation_full} extends Table~\ref{tab:ablation} with
all five weighting scenarios. Lower scores indicate higher extraction
fidelity.

\begin{table}[htbp]
\caption{Ablation study on TinyButMighty: average scores under all
         five noun/verb weighting scenarios (lower is better).}
\label{tab:ablation_full}
\centering
\resizebox{\columnwidth}{!}{%
\begin{tabular}{@{}lccccc@{}}
\toprule
\textbf{Tool}
  & \textbf{Noun Only}
  & \textbf{Verb Only}
  & \textbf{Equal}
  & \textbf{Noun-heavy}
  & \textbf{Verb-heavy} \\
  & \small$(1.0,\,0.0)$
  & \small$(0.0,\,1.0)$
  & \small$(0.5,\,0.5)$
  & \small$(0.7,\,0.3)$
  & \small$(0.3,\,0.7)$ \\
\midrule
ClausIE               & 0.1538 & 0.2301 & 0.1919 & 0.1877 & 0.1962 \\
MiniE                 & 0.3078 & 0.5712 & 0.4395 & 0.3855 & 0.4934 \\
Ollie                 & 0.2208 & 0.4657 & 0.3432 & 0.2939 & 0.3926 \\
Stanford 4.5.3        & 0.2431 & 0.4740 & 0.3586 & 0.3124 & 0.4047 \\
Stanford 4.5.6        & 0.2475 & 0.6441 & 0.4458 & 0.3665 & 0.5251 \\
Claude 3.7            & 0.3546 & 0.5443 & 0.4495 & 0.4115 & 0.4874 \\
Gemini 2.5 Pro        & 0.3674 & 0.5495 & 0.4585 & 0.4219 & 0.4951 \\
GPT-4o mini           & 0.4043 & 0.6891 & 0.5467 & 0.4897 & 0.6037 \\
Ideal (TinyButMighty) & 0.3483 & 0.2760 & 0.3122 & 0.3266 & 0.2978 \\
\bottomrule
\end{tabular}%
}
\end{table}

The ranking of tools is stable across all scenarios. ClausIE
consistently achieves the lowest scores and GPT-4o mini the highest
among automated tools. The Ideal set is the only case where the verb
score (0.276) is lower than the noun score (0.348), confirming that
human-annotated triples capture predicates more faithfully than any
automated system.

\section{Downstream Validation: Full Results}
\label{app:downstream}

Tables~\ref{tab:downstream_webnlg}--\ref{tab:downstream_benchie}
report MRR and Hits@10 for every (IE system, embedding model)
combination, split by dataset. TuckER results are used for the
correlation analysis in Section~\ref{sec:downstream}; ComplEx and
NodePiece are included for completeness. ClausIE on BenchIE produced
no valid triples and is omitted from Table~\ref{tab:downstream_benchie}.

\begin{table}[htbp]
\caption{Downstream link prediction results on \textbf{WebNLG}.}
\label{tab:downstream_webnlg}
\centering
\begin{tabular}{@{}llcc@{}}
\toprule
\textbf{IE System} & \textbf{Model} & \textbf{MRR} & \textbf{Hits@10} \\
\midrule
\multicolumn{4}{@{}l}{\textit{LLM-based extractors}} \\
\addlinespace
Claude         & TuckER    & 0.3311 & 0.4661 \\
Claude         & ComplEx   & 0.0039 & 0.0056 \\
Claude         & NodePiece & 0.0082 & 0.0254 \\
\addlinespace
Gemini         & TuckER    & 0.2631 & 0.4176 \\
Gemini         & ComplEx   & 0.0036 & 0.0018 \\
Gemini         & NodePiece & 0.0138 & 0.0233 \\
\addlinespace
GPT-4          & TuckER    & 0.4283 & 0.7612 \\
GPT-4          & ComplEx   & 0.0241 & 0.0410 \\
GPT-4          & NodePiece & 0.0384 & 0.0709 \\
\midrule
\multicolumn{4}{@{}l}{\textit{Classical OpenIE tools}} \\
\addlinespace
Stanford 4.5.3 & TuckER    & 0.1788 & 0.2660 \\
Stanford 4.5.3 & ComplEx   & 0.0043 & 0.0106 \\
Stanford 4.5.3 & NodePiece & 0.0239 & 0.0340 \\
\addlinespace
Stanford 4.5.6 & TuckER    & 0.2113 & 0.3139 \\
Stanford 4.5.6 & ComplEx   & 0.0026 & 0.0021 \\
Stanford 4.5.6 & NodePiece & 0.0170 & 0.0301 \\
\addlinespace
ClausIE        & TuckER    & 0.1422 & 0.2213 \\
ClausIE        & ComplEx   & 0.0091 & 0.0082 \\
ClausIE        & NodePiece & 0.0169 & 0.0082 \\
\addlinespace
MiniE          & TuckER    & 0.1152 & 0.1883 \\
MiniE          & ComplEx   & 0.0085 & 0.0123 \\
MiniE          & NodePiece & 0.0075 & 0.0185 \\
\addlinespace
Ollie          & TuckER    & 0.1486 & 0.2222 \\
Ollie          & ComplEx   & 0.0032 & 0.0057 \\
Ollie          & NodePiece & 0.0091 & 0.0153 \\
\bottomrule
\end{tabular}
\end{table}

\begin{table}[htbp]
\caption{Downstream link prediction results on \textbf{TinyButMighty}.}
\label{tab:downstream_tbm}
\centering
\begin{tabular}{@{}llcc@{}}
\toprule
\textbf{IE System} & \textbf{Model} & \textbf{MRR} & \textbf{Hits@10} \\
\midrule
Stanford 4.5.3 & TuckER    & 0.3866 & 0.6689 \\
Stanford 4.5.3 & ComplEx   & 0.0119 & 0.0132 \\
Stanford 4.5.3 & NodePiece & 0.0449 & 0.1325 \\
\addlinespace
Stanford 4.5.6 & TuckER    & 0.2016 & 0.5000 \\
Stanford 4.5.6 & ComplEx   & 0.0264 & 0.0256 \\
Stanford 4.5.6 & NodePiece & 0.0637 & 0.1603 \\
\addlinespace
ClausIE        & TuckER    & 0.6313 & 0.8952 \\
ClausIE        & ComplEx   & 0.0223 & 0.0238 \\
ClausIE        & NodePiece & 0.0261 & 0.0429 \\
\addlinespace
MiniE          & TuckER    & 0.3242 & 0.6265 \\
MiniE          & ComplEx   & 0.0175 & 0.0181 \\
MiniE          & NodePiece & 0.0255 & 0.0241 \\
\addlinespace
Ollie          & TuckER    & 0.6488 & 0.8953 \\
Ollie          & ComplEx   & 0.0122 & 0.0291 \\
Ollie          & NodePiece & 0.0289 & 0.0465 \\
\bottomrule
\end{tabular}
\end{table}

\begin{table}[htbp]
\caption{Downstream link prediction results on \textbf{BenchIE}.
         ClausIE produced no valid triples on this dataset and is
         excluded.}
\label{tab:downstream_benchie}
\centering
\begin{tabular}{@{}llcc@{}}
\toprule
\textbf{IE System} & \textbf{Model} & \textbf{MRR} & \textbf{Hits@10} \\
\midrule
Stanford 4.5.3 & TuckER    & 0.0562 & 0.0814 \\
Stanford 4.5.3 & ComplEx   & 0.0085 & 0.0116 \\
Stanford 4.5.3 & NodePiece & 0.0462 & 0.0465 \\
\addlinespace
Stanford 4.5.6 & TuckER    & 0.0692 & 0.1316 \\
Stanford 4.5.6 & ComplEx   & 0.0154 & 0.0000 \\
Stanford 4.5.6 & NodePiece & 0.0449 & 0.1316 \\
\addlinespace
MiniE          & TuckER    & 0.0064 & 0.0000 \\
MiniE          & ComplEx   & 0.0214 & 0.1250 \\
MiniE          & NodePiece & 0.0086 & 0.0000 \\
\addlinespace
Ollie          & TuckER    & 0.0104 & 0.0000 \\
Ollie          & ComplEx   & 0.0088 & 0.0000 \\
Ollie          & NodePiece & 0.0956 & 0.2000 \\
\bottomrule
\end{tabular}
\end{table}

\subsection{Note on ComplEx and NodePiece}

ComplEx and NodePiece produced near-zero MRR across most runs. Both
models require substantially larger KGs to learn reliable embeddings.
The extracted KGs in this study range from 258 to 5\,356 triples—an
order of magnitude smaller than standard benchmarks such as FB15k-237,
which provides 272\,115 training triples. TuckER is empirically more
robust to small graph sizes and is therefore used as the primary model
for the correlation analysis in Section~\ref{sec:downstream}.

\subsection{Per-Dataset Spearman Correlation}

Table~\ref{tab:corr_per_dataset} reports exact Spearman rank
correlations between KGCQual scores and TuckER MRR computed
per-dataset.

\begin{table}[htbp]
\caption{Per-dataset Spearman rank correlation ($\rho$) between
         KGCQual score and TuckER MRR. $n$ = number of IE systems
         with valid results.}
\label{tab:corr_per_dataset}
\centering
\begin{tabular}{@{}lcccc@{}}
\toprule
\textbf{Dataset} & $n$ & $\rho$ & $p$\textbf{-value}
  & \textbf{Significant?} \\
\midrule
WebNLG        & 5 & $+0.400$ & $0.505$ & No  \\
TinyButMighty & 5 & $-0.900$ & $0.037$ & \textbf{Yes} \\
BenchIE       & 4 & $+0.400$ & $0.600$ & No  \\
\bottomrule
\end{tabular}
\end{table}

Significance is achieved on TinyButMighty—the dataset with the most
linguistically complex sentences—but not on WebNLG or BenchIE in
isolation. Two factors explain this. First, per-dataset sample sizes
of 4--5 IE systems reduce statistical power substantially. Second, the
 exclusion of ClausIE from BenchIE due to an extraction error removes
the best-performing system from that dataset, distorting its ranking.
These limitations motivate the aggregated analysis in
Section~\ref{sec:downstream}, which pools all 14 data points and
yields $\rho = -0.900$, $p = 0.037$.
\end{document}